\documentclass[preprint,1p,12pt]{elsarticle}
\usepackage{amssymb}

\usepackage{amsmath}
\usepackage{hyperref}
\usepackage{array}
\usepackage{makecell}
\usepackage{graphicx}
\usepackage{tabularx}
\usepackage{comment}
\usepackage{enumitem}
\usepackage{svg}
\usepackage{booktabs}
\usepackage{xcolor}
\usepackage{xurl}
\usepackage{breakurl}

\journal{Neurocomputing}

\begin{document}

\begin{frontmatter}

\title{A Collaborative Content Moderation Framework for Toxicity Detection based on Multitask Neural Networks and Conformal Estimates of Annotation Disagreement} 

\author[label1]{Guillermo Villate-Castillo} 
\affiliation[label1]{organization={TECNALIA, Basque Research and Technology Alliance (BRTA)},
            postcode={48160}, 
            city={Derio},            
            country={Spain}}
\author[label1,label2]{Javier {Del Ser}} 
\affiliation[label2]{organization={University of the Basque Country (UPV/EHU)},
            postcode={48013}, 
            city={Bilbao},            
            country={Spain}}
\author[label3]{Borja Sanz} 
\affiliation[label3]{organization={Faculty of Engineering, University of Deusto},
            postcode={48007}, 
            city={Bilbao},            
            country={Spain}}

\begin{abstract}

Content moderation typically combines the efforts of human moderators and machine learning models. However, these systems often rely on data where significant disagreement occurs during moderation, reflecting the subjective nature of toxicity perception. Rather than dismissing this disagreement as noise, we interpret it as a valuable signal that highlights the inherent ambiguity of the content—an insight missed when only the majority label is considered. In this work, we introduce a novel content moderation framework that emphasizes the importance of capturing annotation disagreement. In this work, we propose a novel content moderation framework that prioritizes capturing annotation disagreement. Our approach leverages multitask neural networks with transformer architectures as their backbone, where toxicity classification serves as the primary task and annotation disagreement is modelled as an auxiliary task. By framing disagreement as a predictive problem within the multitask learning architecture, our method effectively captures the nuanced ambiguity of content toxicity. Additionally, we leverage uncertainty estimation techniques, specifically Conformal Prediction, to account for the model's inherent uncertainty in predicting toxicity and annotation disagreement. The framework also allows moderators to adjust thresholds for annotation disagreement, offering flexibility in determining when ambiguity should trigger a review. We demonstrate that our joint approach enhances model performance, calibration, and uncertainty estimation, while offering greater parameter efficiency and improving the review process in comparison to single-task methods.

\end{abstract}

\begin{graphicalabstract}
\end{graphicalabstract}

\begin{highlights} 
    \item A novel collaborative Content Moderation framework based on Conformal Prediction.
    \item A multitask neural network with auxiliary  annotation disagreement task to guide reviews.
    \item We propose new metrics, CARE and F1 Review, to measure human collaboration effectiveness.
    \item Experimental results show improvements in uncertainty quantification and calibration.
    \item Multitask approach enhances composite single tasks in content moderation.
\end{highlights}

\begin{keyword}
Conformal prediction \sep Uncertainty quantification \sep Multitask neural network \sep Collaborative content moderation \sep Toxicity detection



\end{keyword}

\end{frontmatter}



\section{Introduction}\label{intro}

Content moderation (CM) has been an important pillar in maintaining ethical online interactions. Given the large amount of user-generated text, CM systems often employ moderation algorithms \cite{gorwa2020algorithmic} to combat the spread of online toxicity. Over the past decade, much research on toxicity detection in text data has leveraged the use of machine learning (ML) models, which have been shown to suffer from reliability and robustness issues, wrong predictions due to spurious lexical features \cite{wang2020identifying} and biases \cite{jiang2020critical}. Robustness and reliability are cornerstones, especially in sensitive areas such as CM, where subjectivity and context significantly influence the results \cite{gorwa2020algorithmic}. 

As observed in a recent survey on toxicity detection \cite{Villate-Castillo2024-tz}, most research contributions reported in this area to date have largely overlooked a key element in decision-making: the estimation of the model's confidence in its prediction. In the few cases when uncertainty quantification (UQ) has been considered \cite{kivlichan-etal-2021-measuring}, models have shown remarkable improvements in robustness. UQ not only optimizes the accuracy of predictions, but also facilitates human moderation. By identifying the least reliable predictions, UQ techniques allow moderators to focus their attention on cases where the model exhibits the highest uncertainty, thus optimizing the use of time and resources devoted to human review.

Incorporating uncertainty has advanced ML models in different disciplines, such as traffic modelling \cite{10472567} or medical imaging \cite{zou2023review}. This advance has been particularly notable in deep neural networks, in which UQ is nowadays crucial for the trustworthiness of such opaque learning models \cite{abdar2021review}. However, it is crucial to recognize that tasks like toxic language detection, which are inherently subjective, require a different perspective due to their complexity. Disagreements among annotators during labelling processes often occur in content moderation reflecting this subjectivity, which can lead to bias and, in some cases, to the censorship of minority opinions. The quantification of inter-annotator disagreement is thus a valuable source of information about the complexity inherent in comments, and is directly related to the aleatoric uncertainty present in the large databases whose inputs require moderation. 

Quantifying annotation disagreement and epistemic uncertainty in the modelling process allows us to identify cases where the model cannot provide robust predictions and must refrain from making them \cite{ARTELT2023126722}. Additionally, it helps determine situations where human moderation is necessary due to complexity or subjectivity. Accounting for both cases is critical, as ML models trained with a majority vote disregard the inherent ambiguity of comments. Unfortunately, considering different annotation perspectives alongside the data modelling pipeline has been observed not to scale well in diverse annotation processes \cite{davani2022dealing}. Additionally, as toxic language evolves over time, it is important to endow content moderators with the tools to make final decisions.

This manuscript builds upon the above observations. Specifically, we propose a novel modeling framework that unifies the primary task of toxic language classification with the auxiliary task of annotation disagreement quantification through a multitask learning architecture approach leveraging a BERT-based Transformer network. The proposed approach is augmented with UQ for both tasks, using Inductive Conformal Prediction (CP) for this purpose. This proposal is grounded on a previous study by \citet{fornaciari2021beyond}, which observed that unifying annotation disagreement and classification can be beneficial for the performance of the primary task. Finally, it aims to address a need highlighted by \citet{gillespie2020content}, who stated: ``\textit{Perhaps, automated tools are best used to identify the bulk of the cases, leaving the less obvious or more controversial identifications to human reviewers, as societies hold together not by reaching perfect consensus, but by keeping their values under constant and legitimate reconsideration}''.

The proposed framework is evaluated over an experimental setup designed to answer with evidence three research questions (RQ):
\begin{itemize}[leftmargin=*]
    \item RQ1: How does the integration of annotation disagreement prediction as an auxiliary task influence the performance and calibration of toxicity detection as the primary task? Conversely, what is the impact of the primary task on the auxiliary task?
    \item RQ2: In what ways does predicting annotation disagreement affect the quantification and interpretation of uncertainty in the toxicity detection task? How does the integration of the primary task influence uncertainty quantification and interpretation in the auxiliary task?
    \item RQ3: What are the benefits, if any, of incorporating an auxiliary task into the toxicity identification process compared to treating them as separate tasks, in terms of robustness and the need for human moderation or revision?
    
\end{itemize}

These three research questions aim to analyse the improvements the multitask approach yields compared to simple single-task models, or, in the case of the full framework, compared to performing quantification separately using a composite of single-task models (CoM).

The rest of the manuscript is organized as follows: Section \ref{background} provides an overview of content moderation and CP on text data, and highlights our main contributions. Section \ref{sec:matmethods} describes the materials and methods employed in our experiments. In Section \ref{experimental_setup}, we outline the experimental setup used to address the RQs. Section \ref{results_discussion} presents the experimental results and includes a discussion of key findings. Finally, Section \ref{conclusion} summarizes the contributions, key insights and limitations, and suggests avenues for future research.

\section{Related Work and Contribution}\label{background}

In this section, we first establish the fundamentals and perform a literature review of CP (Subsection \ref{conformal_prediction}), with a focus on text classification. Furthermore, we review the research area of perspectivism in Section \ref{perspectivism}, with a particular focus on learning from annotation disagreement. Then we define content moderation and revisit frameworks proposed in the recent literature within the crossroads of toxicity detection and uncertainty estimation (Subsection \ref{content_moderation}). Finally, we end the section with a statement on the contribution of our work beyond the literature reviewed in the previous subsections (Subsection \ref{contrib}).

\subsection{Conformal Prediction}\label{conformal_prediction}

CP is a framework pioneered by \citet{vovk2005algorithmic}, which can be defined as ``\textit{a practical measure of the evidence found in support of a prediction, by estimating how unusual a potential example looks with respect to previous ones}'' \citet{gammerman2013learning}. Given a desired confidence level \(1-\alpha\), conformal algorithms are proven to always be valid, meaning that the prediction intervals or sets generated by these algorithms will, on average, correctly contain the true outcomes at least \(1-\alpha\) of the time. This validity is achieved without requiring any specific assumptions about the data distribution, except for the assumption of independent and identically distributed (i.i.d.) data. Conformal predictors can be used for both classification and regression tasks, and can be constructed from any conformity score to indicate the similarity between a new test example and the training examples. Thanks to their model-agnostic nature, CP techniques can be employed alongside any ML algorithm. 

Since we deal with toxicity detection, we depart from a trained classification model (CLASS), from which we create a set of possible predicted labels $\mathcal{C}(\mathbf{x})$ associated to input $\mathbf{x}$ using a small amount of data called calibration data. To construct the prediction set in conformal prediction for classification, we first select a conformal score \( s \), typically derived from the softmax output of the CLASS. We then calculate the quantile \( \widehat{q} = (n+1)(1-\alpha)/n \) (with $n$ denoting the number of calibration samples), which represents a corrected estimate of the \( 1-\alpha \) quantile for our chosen confidence level. Finally, we generate the prediction set $\mathcal{C}(\mathbf{x})$ by including all classes for which the conformal score does not exceed the computed quantile, following the specified conformal prediction method.
 
For the regression models (REG), the description of CP algorithms shifts from generating a prediction set to generating an interval that guarantees coverage of the true value in future observations based on previous ones (i.e., the calibration data). Similar to classification, we need to select a conformity score and compute the corresponding quantile that ensures coverage at the desired confidence level.

Given its robust theoretical foundation and model-agnostic applicability, CP has been widely utilized in various domains, including object detection \cite{moreno2019modelling}, medicine \cite{olsson2022estimating}, weather forecast \cite{wang2020conformal} and natural language processing (NLP). According to a recent survey on the use of CP in NLP \cite{campos2024conformal}, tasks often tackled with CP include binary classification, particularly for paraphrase detection, sentiment analysis, and Boolean question answering \cite{giovannotti2022calibration}; multiclass classification for part-of-speech (POS) tagging \cite{dey2022conformal}; and multilabel classification \cite{maltoudoglou2022well}. In the case of natural language generation, CP has been used at the sentence level to provide guarantees for Large Language Model (LLM) planners to complete tasks \cite{liang2024introspective}, and at the token level to mitigate overconfidence in nucleus sampling text decoding strategies \cite{ulmer2024non}. Although CP has been predominantly applied to classification tasks, we have not observed any instances in the literature where it has been specifically used for toxic language detection, except for the introductory examples provided in \citet{angelopoulos2021gentle}. This lack of research is evident in the next subsection, where we explore content moderation in more detail.

\subsection{Perspectivism: Learning from annotation disagreement}\label{perspectivism}

In supervised learning it is common practice to collect multiple annotator judgments on the same data instances to improve the quality of the final labels \cite{nowak2010reliable}. Although all human judgments carry some degree of subjectivity, certain tasks, including toxicity detection, are widely acknowledged to be particularly subjective \cite{zhang2023taxonomy}. A commonly used approach to fuse different annotations for a given data instance is majority voting, which is widely adopted in the literature when such annotations are available. However, this fusion approach may not always be appropriate. Such aggregation methods often overlook the socially constructed nature of human perceptions, silencing valuable insights especially those from underrepresented groups \cite{prabhakaran2021releasing}. This is particularly problematic because these methods fail to fully capture the diversity of opinions, leading to the loss of important information especially in tasks where a unique ’ground truth’ may not even exist \cite{kumar2021designing}. 

Although annotator disagreement has been identified as a meaningful indicator in these subjective tasks, it is equally important to recognize that such disagreement does not always reflect genuine differences in perception. In many cases, it can arise due to issues in the annotation process, such as sloppy annotations, ambiguity, or missing information, which can negatively impact the quality of these disagreement signals \cite{sandri2023don}. These factors have been widely documented in studies that analyse disagreement as a positive cue, and are commonly considered to be an inherent limitation of these approaches \cite{milkowski2023modeling,mokhberian2023capturing,weerasooriya2023disagreement} due to the difficulty in detecting them \cite{aroyo2015truth}. Given this limitation, \citet{cabitza2023toward} and \citet{aroyo2015truth} have proposed to introduce improved annotation procedures. Key factors highlighted in these works include ensuring a sufficient number of annotators (at least 12 to 15 per data instance) to allow for statistically significant majorities. This recommendation is based on the finding that beyond this number, significant differences in disagreement are rarely observed. In addition, they emphasize the importance of involving a diverse set of raters in terms of origin, culture, background, and other relevant aspects to better capture the complexity of subjective tasks.

The challenge of effectively handling disagreement in annotation leads to the emergence of perspectivism, a theoretical framework and a family of methods in Artificial Intelligence (AI) that takes into account the importance of annotation disagreement for predictive models and embraces the use of labelled data where disaggregated labels are available \cite{frenda2024perspectivist}. There are two categories within perspectivism: weak perspectivism, which focuses on the collection and analysis of disaggregated labels; and strong perspectivism, where disaggregated labels are used for both training and evaluation. In this work we focus on strong perspectivism, particularly in learning from disagreement approaches, where we primarily observe three main methodologies: i) \textit{modelling individual perspectives}, ii) \textit{modelling disagreement among groups of annotators}, and iii) \textit{modelling label distribution}.

To begin with, modelling individual perspectives focuses on learning from each annotator’s labels in isolation, rather than aggregating all annotations under a majority voting rule. All existing approaches in the literature share a common procedure: incorporating an annotator embedding in addition to the text embedding to personalize the model’s predictions. \citet{yin2023annobert} first embraced this methodology. Other approaches, besides including annotator embeddings, also incorporate annotation embeddings as a proxy for modelling rater’s annotation tendencies, as seen in \citet{deng2023you}. In the case of AART \cite{mokhberian2023capturing}, InfoNCE loss is incorporated to bring annotators with similar annotation patterns closer together, thereby improving model performance. Although these approaches have demonstrated better performance compared to single-task models trained on aggregated labels, a key limitation we have observed across all methods modelling individual perspectives is the restricted number of annotations delivered per annotator. Additionally, scalability becomes a challenge when dealing with hundreds or even thousands of annotators.



Secondly, modelling disagreement among groups of annotators is a learning approach that, instead of focusing solely on individual annotators, groups them based on shared characteristics such as religion, gender, political views, or ethnicity, among others. As observed in \citet{fleisig2023majority}, this methodology relies on the idea that individuals with similar backgrounds may share annotation biases that can provide valuable contextual information and improve the predictive performance of individual annotations. One method for incorporating this information is by introducing the characteristics of the annotator as a prior textual prompt to the model’s prediction process using the token separator [SEP], as was done in \citet{fleisig2023majority}. A similar approach is taken by \citet{wan2023everyone},  with an additional step: training the model on labels aggregated within similar annotator groups. A distinct scheme is proposed by \citet{gordon2022jury}, which reframes text classification as a recommendation problem. Their model integrates text embeddings, annotator embeddings, and annotator characteristic embeddings within a cross network and deep network architecture. This structure enables the model to learn personalized classification patterns based on annotator tendencies. However, despite these advantages, these approaches face several challenges. Among them, one major limitation is the difficulty in obtaining personal data characterizing the annotators, as such information is often sensitive or even unavailable. Additionally, individuals who share general demographic traits may still exhibit diverse annotation biases, as personal experiences significantly influence the decisions during the annotation process.

The third category, modelling label distribution, involves capturing the distribution of labels in a disaggregated dataset without explicitly modelling individual or group annotator perspectives. Several methods have been proposed to model label distribution. For instance, \citet{plank2014learning} introduced an inter-annotation agreement loss in the Part-of-Speech (POS) tagging task, leading to improved performance in downstream tasks. Similarly, \citet{xu2024leveraging} reformulated the task as a multilabel classification problem to model the full label distribution provided by annotators.  Other methods consider label distribution as an auxiliary task. For example, \citet{fornaciari2021beyond} predict soft labels, while \citet{sandri2023don} convert the auxiliary task into classifying disagreement levels as none, low, or high. In \citet{rodriguez2024federated}, the authors propose using Federated Learning to jointly train a model across different annotators without explicitly relying on majority voting. This approach enhances robustness and represents more fairly the plurality of opinions. Similarly to the approach proposed in this manuscript, \citet{pavlopoulos2024polarized} treat disagreement as a separate class to be predicted, incorporating the Distance from Unimodality (DFU) metric, which improves predictive performance for other labels. These approaches demonstrate the benefits of leveraging annotation disagreement to enhance the primary task’s performance. However, they face challenges due to the limited number of annotators per instance, making them susceptible to noise from ambiguous, inconsistent, or incomplete annotations.

\color{black}
\subsection{Content Moderation}\label{content_moderation}

CM, as defined in \citet{grimmelmann2015virtues}, refers to the governance mechanisms that structure participation in a community to facilitate cooperation and prevent abuse. In our context, when we talk about CM, we also refer to algorithmic CM, which the same author defines as \textit{``systems that classify user-generated content based on either matching or prediction, leading to a decision and governance outcome''}.

Before the rise of social media, CM was primarily performed by human moderators. However, with the surge of interactions on new platforms, this approach became unfeasible for humans to manage alone. As a result, companies like YouTube, Twitter, and Facebook began to develop their own moderation tools. Other initiatives also emerged, such as Google's Perspective API and OpenAI's CM tools, to support companies that did not have the resources to create their own tools. These algorithmic moderation tools were primarily designed to automate CM, often without fully considering the role of human moderators. However, this system has been observed to have several limitations regarding biases, including issues with state-of-the-art toxicity detectors like the Perspective API, as well as very limited capabilities in detecting deliberately obfuscated toxic sentences \cite{FENZA2024127951}.


Up to this point, we have discussed \emph{automatic} CM, in which there is no interaction between the system and human moderators. In contrast, \emph{collaborative} CM can be defined as a framework in which an algorithmic moderation tool and a human moderator work together to classify user-generated content, each compensating for the other’s weaknesses. While moderation tools excel at processing large volumes of data, they often lack robustness, adaptability, and the ability to contextualize. On the other hand, human moderators can provide broader knowledge, deeper contextual understanding, and empathy.

Building on this definition of collaborative CM and drawing from the literature review in \citet{Villate-Castillo2024-tz}, it becomes evident that very little research has been conducted in this specific area. The few studies that do explore aspects of collaborative CM include those in \citet{wang2024social} and \citet{kivlichan-etal-2021-measuring}. The former study leverages user interactions alongside the knowledge of content moderators to perform classification. Although it introduces the role of moderators and their expertise, it does not address human oversight or governance of the moderation process. In contrast, the latter study is the only one focusing on human oversight. It employs various UQ techniques (though CP was notably excluded) to determine when collaboration between moderators and moderation tools is necessary, particularly in cases where the model’s predictions are uncertain.

\subsection{Main Contribution} \label{contrib}

Taking into account the analysed literature, the main contribution of this manuscript is a collaborative CM framework where the moderator is in control of the algorithmic CM tools by deciding when the moderator’s final decision should be incorporated. This is achieved not only by considering the model’s confidence through UQ, but also by incorporating information about the level of annotation disagreement the model should handle in its predictions. Since toxicity is a subjective task, with opinions varying across content annotators, it is crucial to leave the final judgment to those who are familiar with CM policies in place, as they are the ones who best understand these policies and their context of application. Additionally, given the evolving nature of language, it is of utmost importance to have mechanisms that are aware of this variability, especially in what refers to the ambiguity of textual toxicity. Along this line, the collaborative CM framework proposed in this work introduces information about the annotation disagreement into the modelling process within a multitask learning setting, enriching the conformalized estimation of the model's confidence, and ultimately yielding better and more reliable decisions about the toxicity of the text data fed at its input. Our approach differs from existing annotation disagreement multitask learning methods used for \textit{modeling label distribution} not only by leveraging annotation disagreement to improve the primary task (namely, toxicity classification), but also by transforming the auxiliary task into an interpretable variable that can directly inform correspondingly, the need for revising the comment flagged as toxic. In addition, we explore the incorporation of annotation disagreement within the uncertainty estimation paradigm, enhancing not only the performance but also the confidence calibration of the model. As demonstrated in Section \ref{results_discussion}, our multitask architecture achieves competitive detection performance, better-aligned and well-calibrated confidence estimations, and additional predictive insights (such as annotation ambiguity), enabling more informed moderation decisions for potentially toxic content.

\section{Materials and Methods} \label{sec:matmethods}

In this section, we first introduce the dataset used for the experiments, along with the modifications made to its original version (see Subsection \ref{dataset}). Next, we present the methods employed (Subsection \ref{methods}), where we provide details on model selection, hyperparameter configuration, the computation of annotation disagreement, and the Uncertainty Quantification (UQ) techniques applied during the experiments. Lastly, we conclude by defining the metrics used to validate our proposed research questions and the collaborative Confusion Matrix (CM) framework (Subsection \ref{metrics}).

\subsection{Dataset}\label{dataset}

When working with toxicity modelling datasets, one of the elements that has been recently criticized by the community is the lack of information about the annotation process, such as the number of annotators, their background, the total number of annotations, and the diversity of the annotators \cite{zhang2023taxonomy}. Since disagreement calculation is a central focus of our implementation, it is essential to have a dataset that includes sufficiently diverse opinions and numerous annotators per comment to yield an heterogeneous distribution of annotations.

As exposed in \citet{Villate-Castillo2024-tz}, only two datasets currently meet these requirements: the Jigsaw Unintended Bias in Toxicity Classification dataset \cite{jigsaw-unintended-bias-in-toxicity-classification}, to which the Jigsaw Specialized Rater Pools dataset \cite{goyal2022your} was also added, as the latter included a portion of this dataset with a higher number of annotations and greater diversity; and Wikipedia Detox (WikiDetox) \cite{Wulczyn2017-jm}. However, Wikipedia Detox is not considered in our study because it does not provide the percentage of people who agreed that a given comment is toxic. Furthermore, this latter dataset contains a majority of comments with a low level of disagreement, which may lead to the detection model being overfit to this data subset, rendering it undesirable. Consequently, the model can struggle to learn the lexical features associated with comments that elicit higher levels of disagreement due to its limited exposure to examples from that fraction of the data.

The Jigsaw Unintended Bias in Toxicity Classification dataset, created as part of a challenge organized by Google Jigsaw, provides a high-quality resource that is pre-divided into training, validation, and test subsets. This predefined partitioning enhances the reproducibility of experiments held with this dataset. The dataset contains comments from the archive of the Civil Comments platform, spanning from 2015 to 2017. Compared to other datasets released in the toxicity research area \cite{Villate-Castillo2024-tz}, this dataset includes a significantly larger group of annotators (8,899) in contrast to the smaller groups of tens typically found in other repositories. This large number of annotators offers a substantial advantage, as it provides a more diverse range of opinions across different backgrounds, ethnicities, genders, and ages. Furthermore, each comment is annotated by at least three annotators, with some comments receiving annotations from thousands of annotators. This large number of annotations helps reduce the likelihood of noisy signals resulting from inconsistent labelling, vagueness, or incomplete information and provides a more adequate representation of disagreement, as highlighted in \citet{cabitza2023toward} and \citet{aroyo2015truth}. However, during the dataset preprocessing phase, several caveats were identified:
\begin{itemize}[leftmargin=*]
    \item The dataset contains comments, each annotated by between 3 and several hundred annotators. Since we are targeting a diverse dataset in terms of opinions, we select only the comments that have at least 10 annotators. This yields 315,685 comments in the training set, 38,855 in the test set, and 33,611 in the validation set, out of a total of approximately 2 million comments.
    
    \item Subsection \ref{annotation_computation} shows that annotation disagreement is measured on a scale from 0 to 1, where 0 indicates no disagreement and 1 indicates high disagreement. When applying the formulas explained in this section to the chosen dataset, we observe that the dataset is imbalanced in terms of the level of disagreement. The majority of comments have a disagreement score higher than 0.4, with most falling between 0.4 and 0.8. Despite this imbalance, the dataset includes instances of medium and high disagreement, unlike WikiDetox, which primarily contains disagreement scores between 0 and 0.1.
    
    \item  The original dataset was imbalanced, with 92.07\% non-toxic comments and 7.93\% toxic comments. However, by selecting only comments with at least 10 annotators, the imbalance is reduced to 61.48\% non-toxic comments and 38.52\% toxic comments.
\end{itemize}

Besides adjusting the dataset size to suit the purpose of the study, we perform thorough data cleaning. This process involves removing HTML tags, emojis, accented characters, IP addresses, and contractions. We also deobfuscate swear words to make the comments more easily understandable.

\subsection{Methods}\label{methods}

In this section we introduce the notation, selected neural network architecture, learning objectives, and hyperparameters (Subsection \ref{model_selection}), whereas Subsection \ref{annotation_computation} discusses various methods from the literature for computing disagreement and explain our selected approach. Finally, Subsection \ref{uncertainty_models} presents the CP algorithms considered in our experiments.

\subsubsection{Notation, Neural Network Architecture, Learning Objectives and Hyperparameters}\label{model_selection}

The proposed CM framework is defined as a multitask learning (MTL) task on an annotated dataset \( \mathcal{D} = \{\mathbf{X}, \mathbf{A}, \mathbf{y}, \mathbf{d}\} \), where \( \mathbf{X} \) is a set of text instances, \( \mathbf{A} \) is the set of annotations, \(\mathbf{y} \) is the set of toxic labels, and \( \mathbf{d} \) is the annotation disagreement value. Each entry \( y_i \in \{0, 1\} \) represents the label assigned to \( \mathbf{x}_i \in \mathbf{X} \) based on the majority vote of the annotators in \( \mathbf{a}_i \in \mathbf{A} \); the label that receives the most votes is selected for each instance. Meanwhile, \( d_i \in \mathbb{R}[0, 1] \) represents the computed annotation disagreement for the set of annotations given to the text instance \( \mathbf{x}_i \).
\begin{figure}
    \centering
    \includegraphics[width=\textwidth]{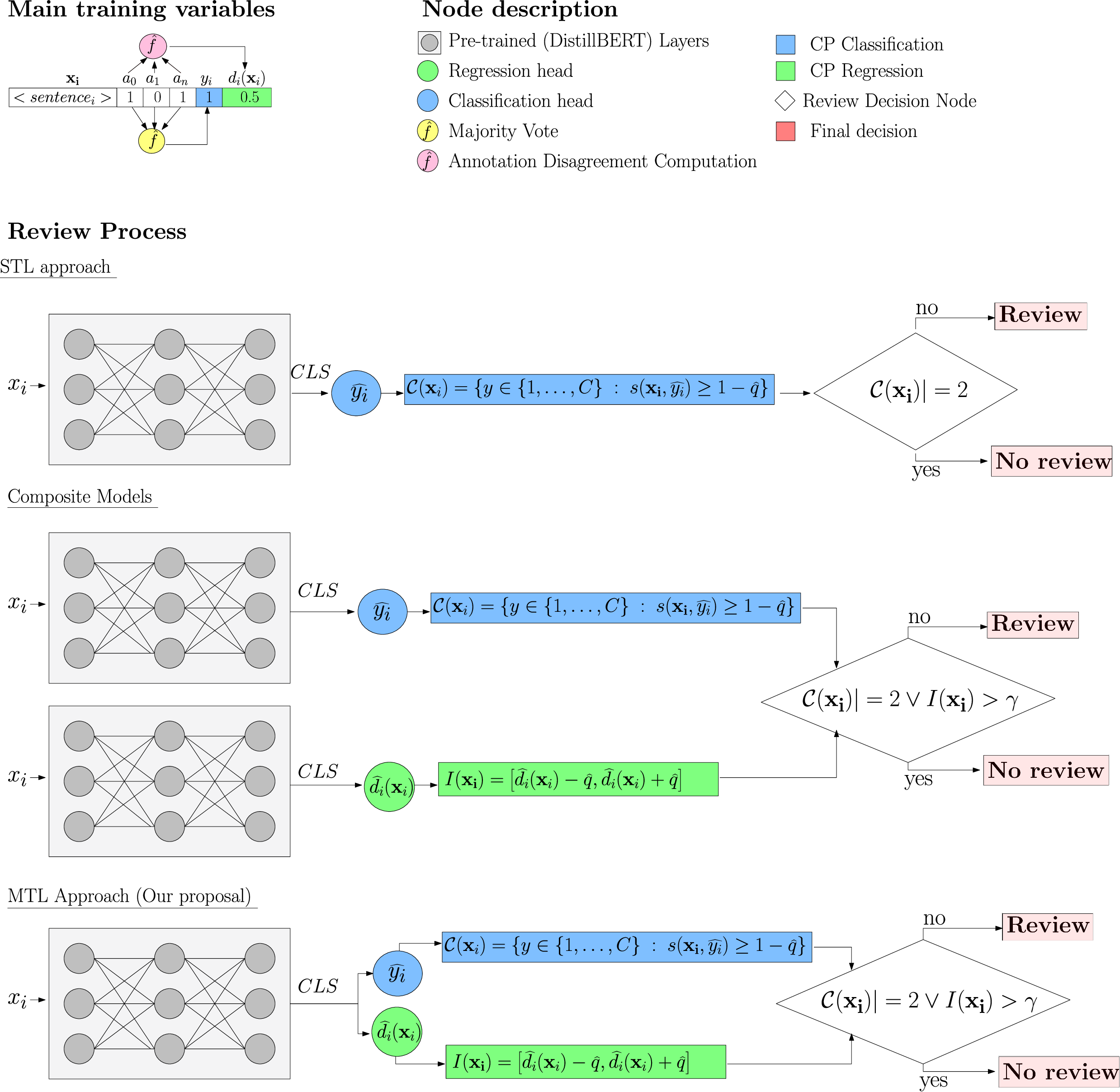}
    \caption{\textbf{Review process.} The STL review process begins by extracting the representation \(\textup{CLS}(\mathbf{x}_i)\) of the content \(\mathbf{x}_i\) and generating its corresponding classification \(\widehat{y}_i\). After calibrating the CP algorithm, we generate the prediction set \(\mathcal{C}(\mathbf{x}_i)\) for classification. If the size of \(\mathcal{C}(\mathbf{x}_i)\) is 2 (i.e., the total number of possible classes in the problem), the comment is flagged for review by a human moderator. Otherwise, the classifier’s output is considered confident, and \(\widehat{y}_i\) is deemed a reliable prediction of the toxicity of \(\mathbf{x}_i\). For the composite STL models, in addition to generating the classification \(\widehat{y}_i\), we also compute the annotation disagreement estimate \(\widehat{d}_i(\mathbf{x}_i)\), a regression value learned from \(\mathbf{X}\) and \(\mathbf{A}\). Once the CP algorithm for regression is calibrated, we produce the confidence interval \(I(\mathbf{x}_i)\). If this interval exceeds a predefined ambiguity threshold \(\gamma\), the comment is marked for human review. Otherwise, we rely on \(\mathcal{C}(\mathbf{x}_i)\) to assess the confidence of the classifier’s output. In the MTL approach, the only difference from the composite model is that both \(\widehat{y}_i\) and \(\widehat{d}_i(\mathbf{x}_i)\) are generated from the same representation \(\textup{CLS}(\mathbf{x}_i)\) of the content.}


    \label{fig:review_process}
\end{figure}

As depicted in Figure \ref{fig:review_process}, the MTL CM process consists of a primary task (toxicity classification of the given comment $\mathbf{x}_i$) and an auxiliary task (estimation of the annotation disagreement). The neural architecture is based on \( \text{DistillBERT}_{\text{base uncased}} \) \cite{sanh2019distilbert}, which serves to extract the embedding from the CLS token (a sentence-level representation for classification) of the given comment \( \mathbf{x}_i \) to perform the MTL task. The generated predictions for each task are defined as \(\widehat{y}_i\) from the primary task and \(\widehat{d}_i(\mathbf{x}_i)\) from the auxiliary task. We then generate the corresponding prediction set \(\mathcal{C}(\mathbf{x}_i) \) and confidence interval \( I(\mathbf{x}_i) \) using the inductive CP techniques later detailed in Subsection \ref{conformal_prediction} and a calibration subset \(\mathcal{D}_{\text{cal}} \). Finally, the need for a human moderator's revision is determined by whether prediction set length of \(\mathcal{C}(\mathbf{x}_i)\) is 2 (namely, the number of possible classes in our problem) or if the predicted confidence interval $I(\mathbf{x}_i)$ is larger than an ambiguity threshold \(\gamma\). The ambiguity threshold \(\gamma\) quantifies the level of disagreement assigned to a comment during the annotation process, where lower values indicate stronger consensus among annotators, and higher values reflect greater divergence in opinions. This criterion allows considering the model's inherent uncertainty in the predictions of the primary and auxiliary tasks, as well as the ambiguity of the comments. By incorporating this threshold, the framework enhances moderation efficiency by reducing the cognitive burden on moderators, allowing them to allocate more attention to complex or novel instances of toxic content.

In the classification task (primary), we address class imbalance, as noted in Subsection \ref{dataset}, by utilizing the so-called \emph{focal loss} \cite{mukhoti2020calibrating} in its weighted form. This loss function has been empirically shown to enhance predictive performance and improve uncertainty calibration in class-imbalanced datasets. For generating the final classification outcome, we employ the Sigmoid activation function, and classify a comment as toxic if the activation output is equal to or above $0.5$.

For the auxiliary regression task, we explore various approaches, including Binary Cross-Entropy (BCE) and Mean Squared Error (MSE) regression losses, as well as treating the regression problem as a classification task. For BCE and MSE, we apply a sigmoid activation function, as the output values are ranged between 0 and 1. When using the regression-as-classification (RAC) approach, we divide the prediction space into 20 equidistant intervals. This method is supported by previous research in \citet{guha2024conformal}, which shows that RAC produces confidence intervals that more accurately reflect the true data distribution.

In line with the recommendations from this study, we utilize their introduced loss function, R2CPP, setting \( \psi = 0.5 \) and \( \tau = 0.1 \). These parameters are maintained constant throughout the experiments discussed later in this paper. Additionally, for all regression loss functions, we implement their weighted versions, applying weights in bins of 0.1 to account for the data imbalance related to annotation disagreement noted in Subsection \ref{dataset}.

Lastly, we introduce single-task learning (STL) models as baselines for both the classification and regression tasks: the single-task regression model (STL REG) and the single-task classification model (STL CLASS). These models are based on the same loss function and hyperparameters used in the MTL framework, aimed to effectively address the research questions introduced in Section \ref{intro}. For RQ1 and RQ2, we utilize the STL models independently to compare the performance, calibration, and uncertainty estimation benefits of our MTL approach for both the primary and auxiliary tasks. Regarding RQ3, we employ CoM for STL REG and STL CLASS during the content review process to assess the actual benefits of the MTL approach in the context of collaborative CM.

\subsubsection{Computation of the Annotation Disagreement}\label{annotation_computation}

Annotation disagreement can be computed in different ways, e.g. as the percentage of people in disagreement with the majority vote as in \citet{wan2023everyone}; by predicting the 5-point scale score as proposed by \citet{ramponi2022dh}; or by using the concept of \emph{soft labels} \cite{fornaciari2021beyond}, i.e., the probability of a comment to be in the positive class. In addition to these methods, we also consider the possibility of using the entropy of the annotations to quantify disagreement. Specifically, given the mean of the annotation labels $\overline{\mathbf{a}}_i$ (which in our case can be regarded as the probability that $\mathbf{x}_i$ is annotated to belong to the \emph{toxic} class), the entropy serves as a measure of uncertainty or unpredictability in the annotation process. This measure is given by:
\begin{equation}
    d(\mathbf{x}_i) = -\overline{\mathbf{a}}_i \log_2(\overline{\mathbf{a}}_i) - (1 - \overline{\mathbf{a}}_i) \log_2(1 - \overline{\mathbf{a}}_i),
\end{equation}
where $\overline{\mathbf{a}}_i$ is the proportion of annotators classifying content $\mathbf{x}_i$ as \emph{toxic}.

In our case we propose an alternative method to compute the annotation disagreement, which draws inspiration from a combination of the works in \citet{wan2023everyone} and \citet{fornaciari2021beyond}. Specifically, the annotation disagreement is given by the distance between the mean of the annotations $\overline{\mathbf{a}}_i$ (assuming that $a_i=1$ represents \emph{toxic} and $a_i=0$ represent \emph{non-toxic}) and $0.5$, which represents the value of maximum disagreement. The resulting value, ranging from 0 to 0.5, is linearly scaled to the range $\mathbb{R}[0, 1]$ and then inverted to yield the sought measure of disagreement. Mathematically:
\begin{equation}
    d(\mathbf{x}_i) = 1 - 2 \cdot \left| \overline{\mathbf{a}}_i - 0.5 \right|,
\end{equation}
where \( d(\mathbf{x}_i) \) represents the annotation disagreement score for \( \mathbf{x}_i\), and $\overline{\mathbf{a}}_i$ is the mean of the annotations associated to $\mathbf{x}_i$.

This metric, the distance based annotation disagreement score, is prioritized over entropy because the latter produces values that are too similar in cases of high uncertainty, and moreover requires a more complex conversion to calculate the percentage of individuals in disagreement. Regarding the scale-based method introduced by \citet{ramponi2022dh}, we were unable to use their proposed technique because the selected dataset lacks the required 5-point scale score for each comment.

\subsubsection{Conformal Algorithms under Consideration}\label{uncertainty_models}

For UQ in classification and regression models, our framework incorporates inductive CP algorithms due to their ability to offer guaranteed coverage, their model-agnostic nature, the fact that they do not require any changes to the model to which they are applied, and because they do not increase latency during inference. To properly validate the advantages offered by using the proposed MTL framework w.r.t. STL baselines, a wide variety of UQ techniques have been selected. For all CP methods, we set \(\alpha=0.1\) to ensure a guaranteed coverage of 90\%.

For classification tasks, the following UQ methods are considered:
\begin{itemize}[leftmargin=*]
    \item \textit{Least Ambiguous set-valued Classifier} (LAC) \cite{sadinle2019least}: In LAC, the conformity score function is defined as one minus the model's softmax probability $p(y_i;\mathbf{x}_i)$ associated to the true label $y_i$, which is computed for each point in the calibration set $\mathcal{D}_{\text{cal}}$. Once the conformity scores \( \{s_1\), ..., \(s_n\} \) are computed for the $n$ calibration instances, we calculate their \((n+1)\dot(1-\alpha)/n\) quantile \(\widehat{q}\). Finally, to compute the uncertainty associated to the samples in the prediction set, for each test instance $\mathbf{x}_i$ we extract the model's predicted class probabilities $p(y;\mathbf{x}_i)$  for each possible class label $y$. All classes with a probability score lower than or equal to the estimated quantile \(\widehat{q}\), i.e.:
    \begin{equation}
    \mathcal{C}(\mathbf{x}_i) = \left\{ y\in\{1,\ldots,C\} : p(y;\mathbf{x}_i) \geq 1 - \widehat{q} \right\},
    \end{equation}
    where $C$ denotes the number of classes in the task under consideration.
    
    \item \textit{Class-Conditional LAC (CCLAC)} \cite{angelopoulos2021gentle}: In this case, we compute the conformity score $s(\mathbf{x},y)$ for each sample $\mathbf{x}$ and class $y$ as one minus the softmax probability assigned by the model to the class, either \textit{toxic} or \textit{non-toxic}. The quantile is then computed for each class, yielding a quantile value $\widehat{q}_y$ for each class $y$ (toxic and not toxic). Finally, during inference we iterate through the classes and include them in the prediction set associated to query instance $\mathbf{x}_i$ if their conformity score $s(\mathbf{x}_i,y)$ is below the corresponding quantile $\widehat{q}_y$, namely:
    \begin{equation}
    \mathcal{C}(\mathbf{x}_i) = \left\{ y\in\{1,\ldots,C\} : p(y;\mathbf{x}_i) \leq \widehat{q}_y \right\},
    \end{equation}
    where, as in LAC, $p(y;\mathbf{x}_i)$ denotes the softmax probability output by the model for class $y$ and input instance $\mathbf{x}_i$. 
    
     \item \textit{Conformal Risk Control (CRC)} \cite{angelopoulos2022conformal}:  It extends the conformal prediction guarantees to any bounded loss function that decreases as $|\mathcal{C}(\mathbf{x}_{n})|$ (i.e. the cardinality of the predicted subset of labels) grows. CRC finds a threshold value $\lambda$ that controls the fraction of missed classes, generating a prediction set $\mathcal{C}_\lambda(\mathbf{x}_i)$ that depends on the selected $\lambda$. In our case, we consider a loss function based on the false negative rate (FNR), which, when computed over the calibration set $\mathcal{D}_{\textup{cal}}$, is given by:
     \begin{equation}
        L^{\text{FNR}}(\lambda) = 1 - \sum_{i=1}^n\frac{|y_i \cap \mathcal{C}_{\lambda}(\mathbf{x}_i)|}{n}, \;
     \end{equation}
      where $\mathcal{C}_{\lambda}(\mathbf{x}_i) = \{y\in\{1,\ldots,C\} : p(y;\mathbf{x}_i) \geq \lambda\}$ is the prediction set given the threshold $\lambda$. This choice ensures that the FNR during inference (namely, when $i>n$) is controlled, i.e. \smash{$E[L^{\textup{FNR}}(\lambda)]\leq \alpha$}.
\end{itemize}

When it comes to the estimation of uncertainty for the auxiliary regression task (annotation disagreement), the following UQ methods are considered:
\begin{itemize}[leftmargin=*]
    \item \textit{Absolute Residual Conformity Score (AR)} \cite{cordier2023flexible}: The conformity score is computed as the residual error between the true value and the predicted value \(| d_i - \widehat{d}_i(\mathbf{x}_i) |\). Then, the quantile of the conformity scores is computed and used to determine the prediction bounds:
    \begin{equation}
    I(\mathbf{x}_i)=\left[\widehat{d}_i(\mathbf{x}_i) - \widehat{q}, \widehat{d}_i(\mathbf{x}_i) + \widehat{q}\right].
    \end{equation}
    
    \item \textit{Gamma Conformity Score (G)} \cite{cordier2023flexible}: The Gscore normalizes the residuals by the predictions as \smash{\(|d_i - \widehat{d}_i(\mathbf{x}_i)|/\widehat{d}_i(\mathbf{x}_i)\)}. As before, we first compute the quantile of the distribution of  normalized conformity scores computed over the calibration set, which is then used to determine the prediction bounds as:
    \begin{equation}
I(\mathbf{x}_i)=\left[\widehat{d}_i(\mathbf{x}_i) \cdot (1 - \widehat{q}), \widehat{d}_i(\mathbf{x}_i) \cdot (1 + \widehat{q})\right].
\end{equation}

    \item \textit{Residual Normalized Conformity Score (RN)} \cite{lei2018distribution}: It is similar to the Gscore. However, instead of dividing by the model's own predictions, this conformal predictor resorts to an external model (K Nearest Neighbors regressor) to predict the REG model residuals \(\widehat{\sigma}(\mathbf{x}_i)\doteq d_i-\widehat{d}_i(\mathbf{x}_i)\). As before, the quantile is computed from the conformity scores, which is then used to determine the prediction bounds:
    \begin{equation}
    I(\mathbf{x}_i)=\left[\widehat{d}_i(\mathbf{x}_i) - \widehat{q} \cdot \widehat{\sigma}(\mathbf{x}_i), \widehat{d}_i(\mathbf{x}_i) + \widehat{q} \cdot \widehat{\sigma}(\mathbf{x}_i)\right].
    \end{equation}
    
    \item \textit{Regression as Classification (R2CCP)} \cite{guha2024conformal}: It is a very recent CP method that converts a regression problem into a classification problem, allowing CP methods devised for classification tasks to elicit conformal intervals for regression. It introduces a new loss function, to preserve the ordering of the continuous alphabet of the target variable. First, the continuous target variable is divided into discrete bins $b\in\{1,\ldots,B\}$. After binning, each instance in the dataset is assigned a class label corresponding to the bin it belongs to. R2CCP's loss can be defined as:
    \begin{equation}
        L^{R2CCP}(\mathbf{x_i})= |d(\mathbf{x_i})- d(b)|^\psi\cdot p(d(b);\mathbf{x_i})- \tau \cdot H(p(d(\mathbf{x_i});\mathbf{x_i})),
    \end{equation}
    where \( d(b) \) denotes the center of the bin \( b \in \{1, \ldots, B\} \); \( |d(\mathbf{x}_i) - d(b)| \) represents the distance between the real disagreement value and the center of bin \( b \); \( \psi>0 \) is a hyperparameter; \( p(d(b); \mathbf{x}_i) \) is the probability of the comment \( \mathbf{x}_i \) belonging to bin \( b \); \( \tau > 0 \) is a regularization term; and \( H(p(d(\mathbf{x}_i); \mathbf{x}_i)) \) is the Shannon entropy of the discrete probability distribution over the bins of the input \( \mathbf{x}_i \).
    
    To compute the quantile \(\widehat{q}\), R2CCP uses a linear interpolation of the softmax probabilities
    as the conformity score $s(\mathbf{x},d(b))$. Finally, the final prediction interval for the continuous variable $d_i$ is the union of the bins within the predicted conformal set of bins, i.e.:
    \begin{align}
    I(\mathbf{x}_i)= \left[d\left(\min \left\{b \in \{1,\ldots,B\} :\; s(\mathbf{x}_i,d(b)) \geq \widehat{q} \right\}\right)\right.,\nonumber \\
    \left.d\left(\max \left\{ b\in\{1,\ldots,B\} :\; s(\mathbf{x}_i,d(b)) \geq \widehat{q} \right\}\right)\right].
    \end{align}
\end{itemize}

\subsection{Performance Metrics}\label{metrics}

Traditional CM systems have been evaluated using performance metrics usually considered in Machine Learning, including accuracy, F1 score, and AUC (\textit{Area under the ROC Curve}). However, these metrics only capture model performance and do not account for the efficiency of models in a collaborative environment with human moderators. 

To overcome this, we henceforth define collaboration efficiency as the model's ability to express its own limitations. In \citet{kivlichan-etal-2021-measuring}, this gap in evaluation metrics is addressed by introducing Review Efficiency, which we rename in our study as \emph{Model Uncertainty-aware Review Efficiency} (MURE) to distinguish between review efficiency based on model uncertainty and annotation disagreement. MURE quantifies the proportion of examples delivered to a moderator in situations where the model’s toxicity classification would otherwise be incorrect. In simpler terms, it measures how often comments flagged as uncertain by CP methods correspond to cases where the model would have made an incorrect prediction.

We mathematically define this metric in the context of a classification task comprising $C$ classes, such that \(\left|\mathcal{C}(\mathbf{x_i})\right| > 1 \) denotes the case when the classifier is not certain about its prediction. We define \(\textup{TP}_{\textup{UQ}}\) as the number of instances where the model correctly delivers the content at its output to the moderator based on its estimated uncertainty, i.e. when $y_i \neq \widehat{y}_i$ and $\left|\mathcal{C}(\mathbf{x}_i)\right| > 1$. Conversely, $\textup{FP}_{\textup{UQ}}$ refers to the number of examples where the model incorrectly sends the content for review based on their estimated uncertainty despite its toxicity prediction being accurate, corr. $y_i = \widehat{y}_i$ and $\left|\mathcal{C}(\mathbf{x}_i)\right| > 1$. As a result, MURE can be computed as:
\begin{equation}
    \textup{MURE} = \frac{\textup{TP}_{\textup{UQ}}}{\textup{TP}_{\textup{UQ}} + \textup{FP}_{\textup{UQ}}}.
\end{equation}

Likewise, to account for the efficiency in predicting when a comment is annotated ambiguously, we propose a new metric coined as \textit{Comment Ambiguity-aware Review Efficiency} (CARE). CARE evaluates the ability to identify all ambiguous comments, based on a threshold \(\gamma\) selected by the moderator. Let $\textup{TP}_{\Leftrightarrow}$ be the number of comments that are truly ambiguous as per $\gamma$ (i.e., \( d(\mathbf{x}_i) \geq \gamma \)) and which are correctly predicted as ambiguous (i.e., \(I(\mathbf{x}_i) \geq \gamma \)). Likewise, we define $\textup{FN}_{\Leftrightarrow}$ as the number of comments which are truly ambiguous but are not predicted as such, i.e., $d(\mathbf{x}_i) \geq \gamma$ but \(  I(\mathbf{x}_i) < \gamma \). Based on these definitions, CARE is given by:
\begin{equation}
\text{CARE} = \frac{\textup{TP}_{\Leftrightarrow}}{\textup{TP}_{\Leftrightarrow} + \textup{FN}_{\Leftrightarrow}},
\end{equation}
namely, the True Positive Rate (TPR) associated to the conformalized annotation disagreement estimate resulting from the auxiliary modelling task.

In addition to CARE and MURE, we also consider \textit{Review F1 Score} (R-F1), which measures how well the framework manages model errors due to both model's uncertainty and comment ambiguity. R-F1 combines precision (MURE) and recall (CARE) to evaluate the system's effectiveness in handling these errors, providing a balance between the two:
\begin{equation}
    \text{R-F1} = 2 \cdot \frac{\textup{MURE} \cdot \textup{CARE}}{\textup{MURE} + \textup{CARE}}.
\end{equation}

Furthermore, we introduce the point-biserial correlation \( r_{pb} \) to capture the correlation between the auxiliary task flagging a comment as ambiguous given the threshold \( \gamma \) and the level of annotation disagreement for that comment, as well as the correlation between the UQ method flagging the primary task prediction as uncertain and the annotation disagreement. As mentioned before, a comment can be deemed ambiguous if \( I(\mathbf{x}_i) \geq \gamma \) and the primary task prediction can be deemed uncertain if \( \left| \mathcal{C}(\mathbf{x}_i) \right| > 1 \). The point-biserial correlation provides us with a score ranging from -1 to 1, where -1 indicates a negative correlation, 1 indicates a positive correlation, and 0 indicates no correlation. This score allows us to analyse the correlation between a binary variable—whether the comment is ambiguous or the prediction is uncertain—and a continuous variable, the annotation disagreement. Keeping this in mind, the point-biserial correlation \( r_{pb} \) can be expressed as:

\begin{equation}
     r_{pb} =\frac {\overline{d_{0}}(\mathbf{x})-\overline{d_{1}}(\mathbf{x})}{\sigma(d(\mathbf{x}))}  \sqrt {\frac {|d_{1}(\mathbf{x})| \cdot |d_{0}(\mathbf{x})|}{|d(\mathbf{x})|^ {2}}}  
\end{equation}
where \( d_1(\mathbf{x}) \) is the group of comments with their corresponding true annotation disagreement values that are ambiguous or for which the model is uncertain in each prediction, and \( d_0(\mathbf{x}) \) is the group of comments that are not ambiguous and for which the model does not show uncertainty, along with their corresponding true annotation disagreement values. \( |d_{1}(\mathbf{x})| \) denotes the number of elements in the ambiguous or uncertain group. \( \sigma(d(\mathbf{x})) \) represents the standard deviation of the annotation disagreement, and \( |d(\mathbf{x})|^{2} \) denotes the total number of disagreement scores elevated to the power of 2.

The metrics presented so far measure how well the framework performs in a CM context. However, we have also introduced additional metrics to support the analysis of each research question. In RQ1, where we analyse the general performance and calibration of the models, we use F1 score, logistic loss, Expected Calibration Error (ECE), and Adaptive Calibration Error (ACE) \cite{nixon2019measuring} for classification, and Mean Absolute Error (MAE) and Mean Squared Error (MSE) for regression. For RQ2, where we measure model uncertainty in classification and regression, the following metrics are used:
\begin{itemize}[leftmargin=*] 
    \item \textit{Classification}: In this case we consider Marginal Coverage, MURE, certainty F1 score, and correlation. Marginal Coverage represents the probability that the prediction set contains the true label, which should ideally approach \(1-\alpha\). The certainty F1 score is the F1 score calculated for the set of comments where the CP method is certain, i.e., \(\left|\mathcal{C}(\mathbf{x}_i)\right| = 1\). Correlation refers to the point-biserial correlation between the model’s uncertainty and annotation disagreement. 
    
    \item \textit{Regression}: In this second case we report mean interval size, Interval Coverage Probability (ICP), distance to interval (DI), and correlation. ICP is the probability that the true prediction lies within the predicted interval, which should also approach \(1-\alpha\). DI measures the mean distance from the interval when the true value $d_i$ falls outside the estimated confidence interval $I(\mathbf{x}_i)$. Correlation, in this case, refers to the Pearson correlation \(r_{I,d}\) between the size of the estimated interval  $I(\mathbf{x}_i)$ and the true annotation disagreement \(d(\mathbf{x}_i)\).
\end{itemize}

Table \ref{tab:metric_resume} summarizes the metrics used to evaluate the classification and regression tasks considered in our framework when addressing each of the formulated RQs.
\begin{table}[h!]    
    \centering
    \resizebox{0.9\textwidth}{!}{%
    \begin{tabular}{cccc}
        \toprule
        \textbf{Task} & \textbf{RQ1} & \textbf{RQ2} & \textbf{RQ3}\\
        \midrule
        \makecell{Content toxicity\\detection\\(primary, classification)} & \makecell{F1 score,\\ Logistic loss,\\ ECE, ACE} & \makecell{Marginal Coverage,\\ MURE,\\ prediction length,\\ certainty F1 score,\\\( r_{pb}\)} & R-F1\\
        \midrule
        \makecell{Annotation ambiguity\\estimation\\(auxiliary, regression)} & MAE, MSE & \makecell{Mean interval size,\\ ICP, DI, \(r_{I,d}\)} &\makecell{R-F1, CARE, \(r_{I,d}\)}\\
        \bottomrule
    \end{tabular}%
    }
\caption{Summary of the metrics considered for each research question RQ.}    
    \label{tab:metric_resume}
\end{table}

\section{Experimental Setup}\label{experimental_setup}

In this section, we describe the experiments conducted to address the research questions outlined in Section \ref{intro}. Section \ref{rq1} and \ref{rq2} detail the experimental setup for RQ1 and RQ2, respectively. Finally, Section \ref{rq3} describes the experimental setup for RQ3.

For all experiments the same hyperparameter setting is used, and fixed seeds are set for the sake of reproducibility. As introduced in Section\ref{methods}, we use the \( \text{DistillBERT}_{\text{base uncased}} \) model, which was fine-tuned on a H100 GPU with a batch size of 32, a maximum sequence length of 512, a learning rate equal to \(2 \cdot 10^{-5}\), a weight decay of $0.01$, and a learning rate warm-up of 1 epoch. AdamW is used as the optimizer, and the model is trained until no improvement is observed in the loss metric, with a patience parameter of 2 epochs. All scripts and log files used to produce the results discussed in the paper are publicly available in  \url{https://github.com/TheMrguiller/Collaborative-Content-Moderation}.

\subsection{RQ1: Impact of Annotation Disagreement Prediction on Primary Task Performance and Calibration}\label{rq1}

The STL CLASS model, depicted in the upper section of Figure \ref{fig:review_process}, is trained using \emph{focal loss} on hard labels. In contrast, the MTL approach, illustrated in the lower section of the same figure, applies the same loss function but incorporates an auxiliary task. To evaluate potential improvements, we compare the STL CLASS across various regression settings, which include BCE, MSE and RAC, as detailed in Subsection \ref{methods}. Given that previous research observed improvements in classification \cite{fornaciari2021beyond}, we hypothesize that a similar effect might occur with the regression auxiliary task. To verify this, we compare each multitask model with the STL REG using the corresponding regression loss. We use MAE and MSE as the primary metrics to evaluate improvements.

These comprehensive assessments are crucial for addressing subsequent research questions related to model uncertainty, particularly because calibration plays a significant role in CP methods. A proper calibration ensures that the estimated prediction intervals or conformal sets provided by CP methods accurately reflect the true uncertainty of the model's predictions.

\subsection{RQ2: Effects of Annotation Disagreement Integration on Uncertainty Quantification }\label{rq2}

Since proper model calibration is closely related to accurate UQ provided by CP methods, we hypothesize that multitask settings could improve model uncertainty estimation for both STL classification (CLASS) and regression (REG). To quantify improvements in the primary task (STL CLASS), we focus specifically on the MURE metric, which assesses how well the uncertainty method predicts cases where the model is incorrect. In addition to this metric, we also consider measures commonly used to evaluate uncertainty estimation methods, including marginal coverage, the correlation between uncertainty and annotation disagreement scores \(r_{I,d}\) (which refers to the Pearson correlation between the estimated annotation disagreement based on the confidence interval and the true annotation disagreement), and the certainty F1 score. To compare uncertainty improvements, we use the same classification and multitask models described in Section \ref{rq1}, calibrated using the same validation/calibration dataset and evaluated on the same test set. Each model is calibrated with the three uncertainty estimation methods introduced in Section \ref{methods}: LAC, Class Conditional LAC, and CRC.

For the auxiliary regression task, a similar approach is used, with the uncertainty metrics focusing on general regression uncertainty quantification metrics, such as interval size, ICP, correlation between interval size and annotation disagreement scores, and DI. Regarding UQ techniques, different methods are employed depending on the type of regression loss. For models based on MSE and BCE regression losses, we use \textit{AR Score}, \textit{G Score}, and \textit{RN Score}. Lastly, for RAC, the previous methods are applied together with \textit{R2CCP}.

These experiments are crucial for assessing the impact of inaccurate model uncertainty quantification on collaborative CM. Misestimation of uncertainty could increase the moderators' workload. For instance, comments that are relatively easy to classify, such as those containing slurs or swear words \cite{anand2024don}, might be unnecessarily submitted to them for revision.

\subsection{RQ3: Benefits of an Auxiliary Task versus Independent Models in Collaborative Content Moderation}\label{rq3}

Previous studies  have assessed the efficiency of moderators and moderation systems using review efficiency (MURE) \cite{kivlichan-etal-2021-measuring}, which considers only model performance while disregarding label data quality and text ambiguity. To address these factors, our moderation framework includes an auxiliary task designed to measure this concept. Since this is a novel approach, we introduce a new metric, CARE, to account for the system's capacity to detect ambiguous examples, as explained in Section \ref{metrics}. To determine if a comment is ambiguous, a threshold \(\gamma\) must be selected that aligns with the desired maximum level of annotation ambiguity. Given that the quantity of uncertain examples varies depending on each model's capabilities, which in turn affects the CARE metric, we used the False Positive Rate (FPR) at a given TPR/MURE value, a metric commonly used in other areas including Out-of-Distribution (OoD) detection \cite{henriksson2021performance}. For BCE and MSE, we apply a TPR of 85\%, while for RAC, we use a TPR of 95\%. Different TPR values are applied for RAC models due to the interval size, which affects the quality of the ambiguity estimation. These values were selected to provide a fair and consistent basis for comparing the performance of each approach. By choosing TPR values that balance precision and recall for each model, we ensure that we can effectively measure and compare each method’s ability to handle uncertainty across different models.

Experiments for RQ3 consider the following models: 
\begin{itemize}[leftmargin=*] 
    \item STL CLASS (upper section of Figure \ref{fig:review_process}): In this case, neural network's uncertainty is used to measure how well the network performs in detecting ambiguous terms, as uncertainty is sometimes associated with hard examples—instances that the model has not learned well. Since the model is not calibrated for the task, we could not compare it under the same true positive rate (TPR) because it is only able to achieve a maximum TPR of around 45\%.
    \item CoM (middle section of Figure \ref{fig:review_process}): To analyse whether there is any additional advantage compared to those examined in RQ1 and RQ2, we use STL CLASS and REG jointly to provide classification and regression uncertainty values. The same procedure employed in the multitask setting is applied to compute the ambiguity of the comment.
    \item MTL (lower section of Figure \ref{fig:review_process}): We use the auxiliary task's confidence interval obtained from RQ2 to quantify the ambiguity of comments based on the threshold derived from computing the FPR at TPR. To determine if a comment is ambiguous, we check whether the upper confidence interval exceeds the selected threshold. 
\end{itemize}

In addition to computing the CARE metric, we also calculate the Review F1 score. Finally, as in previous experiments, we compute the correlation between the ambiguity score predicted by the model and the true annotation disagreement.

\section{Results and Discussion}\label{results_discussion}

In this section, we present and analyse the results for each research question. In Section \ref{rq1_results} we discuss our findings for RQ1, highlighting the benefits of our approach in terms of model calibration and performance. Section \ref{rq2_results} focuses on the advantages of incorporating the auxiliary task for improving uncertainty estimation in the primary classification task. In Section \ref{rq3_results} we compare the multitask and composite approaches for content moderation, with a focus on comment annotation disagreement. For each RQ, we begin by summarizing the key performance metrics and findings.

\subsection{RQ1: Impact of Annotation Disagreement Prediction on Primary Task Performance and Calibration}\label{rq1_results}

Following the experimental setup presented in Section \ref{rq1}, Table \ref{tab:performance_classification_weighted} summarizes the performance results for the primary task (toxicity detection) of the STL CLASS approach and the MTL approach with different regression settings. By examining the F1 score for the classification task, it is evident that the introduction of the auxiliary task improves the toxicity detection performance, especially in the cases of MSE and BCE, and to a lesser extent, RAC. In terms of calibration, the BCE and MSE MTL models exhibit better calibration and, consequently, achieve higher F1 scores compared to their STL CLASS counterpart. However, for RAC MTL, we do not observe any significant improvement in calibration as per the ACE and logistic loss values when compared to the STL CLASS approach, despite RAC providing more information about the distribution of annotation disagreement scores (which closely relates to the toxicity score).
\begin{table}[h!]
    \centering
    \begin{tabular}{ccccc}
        \toprule
        Model name & F1 score \(\uparrow\) & ECE \(\downarrow\) & ACE \(\downarrow\) & Log loss \(\downarrow\) \\
        \midrule
         STL CLASS & 71.61 & 0.0960 & \(1.039 \times 10^{-4}\) & 0.6407 \\
        
        MTL + BCE & 72.15 & \underline{0.12} & \(1.013 \times 10^{-4}\) & \textbf{0.6326} \\
        
        MTL + MSE & \textbf{72.48} & \underline{0.12} & \(\textbf{1.002} \times 10^{-4}\) & 0.6372 \\
        
        MTL + RAC & 71.89 & 0.0726 & \underline{\(1.058 \times 10^{-4}\)} & \underline{0.6570} \\
        \bottomrule
    \end{tabular}%
    \caption{Comparison of detection performance and calibration between STL CLASS and MTL models for toxicity classification. The best model for each metric is in bold, and models with worse metrics than the baseline are underlined.}
    \label{tab:performance_classification_weighted}
\end{table}

Surprisingly, ECE does not show consistency with other calibration metrics, which supports the findings of \citet{nixon2019measuring}, who observed that ECE is not a reliable metric.

In the case of STL for the regression task (STL REG), Table \ref{tab:performance_regression} exposes a clear improvement in all cases, as indicated by the MAE and MSE metrics. The largest improvement compared to the STL REG model occurs in RAC, where the classification task further supports the regression task. The converse does not hold for the classification task.
\begin{table}[h!]
    \centering
    \begin{tabular}{ccccc}
        \toprule
        Model name & \makecell{MSE \(\downarrow\)\\ STL REG/ MTL} & \makecell{MAE \(\downarrow\)\\ STL REG/ MTL}\\
        \midrule
        
        BCE & 0.0461/\textbf{0.0444} & 0.1694/\textbf{0.1662} \\

        MSE & 0.0460/\textbf{0.0458} & 0.1693/\textbf{0.1688} \\

        RAC & 0.0673/\textbf{0.0521} &  0.2075/\textbf{0.1831 }\\
        \bottomrule
    \end{tabular}
    \caption{Comparison of performance between STL REG and MTL models for regression tasks. The best model for each metric is in bold, and models with worse metrics than the baseline are underlined.}
    \label{tab:performance_regression}
\end{table}

In summary, we observe a clear improvement in the classification task for the MSE and BCE multitask versions compared to the STL CLASS, with RAC version showing a poorer calibration. In STL REG, there is a more consistent overall improvement for the MSE and MAE metrics. Although the improvements in classification are not substantial, our results are in line with the findings of previous studies \cite{fornaciari2021beyond,sandri2023don}.

\subsection{RQ2: Effects of Annotation Disagreement Integration on Uncertainty Quantification}\label{rq2_results}

As described in the experimental setup corresponding to RQ2, we now focus on analysing whether there is any clear improvement in uncertainty quantification for the regression and classification tasks by including an auxiliary task in annotation disagreement prediction.

The results for the classification uncertainty are divided based on the conformal prediction methods in use:
\begin{itemize}[leftmargin=*] 
    \item LAC: Table \ref{tab:basecp_weighted} presents a comparison between the STL CLASS 
    and its MTL counterparts. For the MURE metric, all MTL models show a clear improvement over the STL CLASS, with the most substantial gains observed in the RAC regression-based models. The F1 score also exhibits a noticeable improvement across all MTL baselines. Furthermore, the higher point-biserial correlation \(r_{pb}\) achieved by the MTL models indicate a stronger alignment with the annotation disagreement, and thus a more effective contextualization of uncertainty.
    
    \item Class Conditional LAC: In Table \ref{tab:classcp_weighted}, we compare the STL CLASS models with the MTL architecture for the Class Conditional LAC uncertainty estimation method. Similar to previous cases, we observe improvements across all metrics compared to STL CLASS, except for the F1 score in the case of R2CCP's regression loss, which performs slightly worse than the STL CLASS model.

    \item CRC: Table \ref{tab:conformalcp_weighted}  reports the results of a comparison between STL CLASS with the MTL architecture for CRC. The results here are consistent with those of LAC, leading to similar conclusions.
\end{itemize}

When it comes to the results of the regression uncertainty quantification, we have divided the analysis into the STL REG models and regression uncertainty quantification methods:
\begin{itemize}[leftmargin=*]
    \item BCE: In Table \ref{tab:regression_cp_1}, we observe that the addition of a classification head leads to better results in terms of interval size, while maintaining the same distance to interval values across all uncertainty methods. However, we generally see worse results in terms of \(r_{I,d}\). The method with the best \(r_{I,d}\) value is G (\emph{Gamma Conformity Score}, as per Section \ref{uncertainty_models}), which generates intervals that are more aligned with the inherent annotation complexity of the comments.

    \item MSE: In Table \ref{tab:regression_cp_1} we note slightly worse results in terms of interval size than in the previous case. However, we obtain smaller values for the distance to the interval, likely due to the interval size being wider than in the BCE case. In this case, the \(r_{I,d}\) metric shows marginally better results.

    \item RAC: In Table \ref{tab:regression_cp_1}, we find that results are consistent with those encountered in BCE. In terms of interval size, there is an overall improvement over those corresponding to STL REG. For the distance to interval metric, we observe smaller values in all cases except for the R2CCP uncertainty quantification method. In terms of \(r_{I,d}\), improvements are observed, except for the AR uncertainty method, where the STL REG model improves upon the multitask model.
\end{itemize}

In summary, we observe improvements in classification across all uncertainty estimation methods compared to the STL models. The model shows better alignment between uncertainty quantification and annotation disagreements, indicating improved handling of difficult examples. However, in regression, the \(r_{I,d}\) metric lacks consistent improvement, depending on the uncertainty method: AR is less aligned, while RN, G, and RC2CPP are more aligned. This may be due to AR providing static intervals, while the others adjust for each input.

For interval size and distance, RAC and BCE losses outperform STL REG models, while MSE yields inconsistent results. We posit that the effectiveness of the auxiliary task relies on the chosen regression loss, with less pronounced improvements in regression due to its sensitivity to annotation disagreements.
\begin{table}
    \centering
    \begin{tabular}{ccccc}
        \toprule
        Model name & MURE \(\uparrow\) & F1 score \(\uparrow\) & Marginal Coverage & \( r_{pb}\) \(\uparrow\)\\
        \midrule
        STL CLASS & 0.4076 & 0.8046 & 0.901 & 0.2767 \\
        MTL + BCE & 0.4084 & 0.8068 & 0.90 & \textbf{0.2830} \\
        MTL + MSE & 0.4105 & \textbf{0.8123} & 0.898 & 0.2775 \\
        MTL + RAC & \textbf{0.4196} & 0.8094 & 0.898 & 0.2770 \\
        \bottomrule
    \end{tabular}
    \caption{Comparison of LAC uncertainty method performance between STL CLASS and MTL models for classification tasks. The best model for each metric is in bold, while models with worse metrics than the baseline are underlined.}
    \label{tab:basecp_weighted}
\end{table}

\begin{table}[h]
    \centering
    \begin{tabular}{ccccc}
        \toprule
        Model name & MURE \(\uparrow\) & F1 score \(\uparrow\) & Marginal Coverage & \( r_{pb}\) \(\uparrow\)\\
        \midrule
        STL CLASS & 0.3528 & 0.8286 & 0.90 & 0.2424 \\
        
        MTL + BCE & 0.3532 & \textbf{0.8304} & 0.8994 & 0.2484 \\
        
        MTL + MSE & 0.3635 & 0.8294 & 0.8987 & \textbf{0.2495} \\
        
        MTL + RAC & \textbf{0.3650} & \underline{0.8251} & 0.8969 & 0.2476 \\
        \bottomrule
    \end{tabular}
    \caption{Comparison of Class Conditional LAC uncertainty method performance between STL CLASS and MTL models for classification tasks. The best model for each metric is in bold, and models with worse metrics than the baseline are underlined.}
    \label{tab:classcp_weighted}
\end{table}
\begin{table}[h]
    \centering
    \resizebox{\textwidth}{!}{ 
    \begin{tabular}{ccccc}
        \toprule
        Model name & MURE \(\uparrow\) & F1 score \(\uparrow\) & Marginal Coverage & \( r_{pb}\) \(\uparrow\)\\
        \midrule
        STL CLASS & 0.4078 & 0.8045 & 0.901 & 0.2763 \\
        
        MTL +  BCE & 0.4084 & 0.806 & 0.8981 & \textbf{0.2830} \\
        
         MTL +  MSE & 0.4105 & \textbf{0.8123} & 0.8989 & 0.2775 \\
        
         MTL +  RAC & \textbf{0.4196} & 0.8094 & 0.8986 & 0.2770 \\
        \bottomrule
    \end{tabular}
    } 
    \caption{Comparison of CRC uncertainty method performance between STL CLASS and MTL models for classification tasks. The best model for each metric is in bold, while models with worse metrics than the baseline are underlined.}
    \label{tab:conformalcp_weighted}
\end{table}

\begin{table}
    \centering
    \resizebox{\textwidth}{!}{ 
    \begin{tabular}{cccccc}
        \toprule
        Model name & Method CP & Interval size \(\downarrow\) & DI \(\downarrow\) & ICP & \(r_{I,d}\) \(\uparrow\)\\
        \midrule
        
        STL REG  BCE & AR & 0.6437 & 0.0096 & 0.8994 & -0.0320\\
         MTL  BCE &  & \textbf{0.6339} & \textbf{0.0094} & 0.8997 & \underline{-0.0910}\\
        \midrule

        STL REG  BCE & G& 0.7198 & 0.0123 & 0.8968 & 0.4963\\
         MTL  BCE &  & \textbf{0.7018} & \textbf{0.0114} & 0.8979 & \textbf{0.5041}\\
        \midrule

        STL REG  BCE & RN & 0.7347 & 0.0113 & 0.8977 & 0.0065\\
         MTL  BCE &  & \textbf{0.7276} & 0.0113 & 0.8984 & \underline{-0.0082}\\
        \midrule

        STL REG  MSE & AR & 0.6424 & 0.0096 & 0.8990 & -0.0357\\
         MTL  MSE &  & \textbf{0.6408} & \underline{0.0098} & 0.8996 & \underline{-0.0600}\\
        \midrule

        STL REG  MSE & G& 0.7166 & 0.0123 & 0.8964 & 0.4959\\
         MTL  MSE &  & \underline{0.7246} & \underline{0.0124} & 0.8985 & \textbf{0.5047}\\
        \midrule
        STL REG  MSE & RN & 0.7320 & 0.0117 & 0.8950 & 0.0000\\
         MTL  MSE &  & \underline{0.7338} & \textbf{0.0112} & 0.8990 & \textbf{0.0010}\\
        \midrule

        STL REG  RAC & AR & 0.72 & 0.0140 & 0.9149 & 0.3659\\
         MTL  RAC &  & \textbf{0.7053} & \textbf{0.0081} & 0.9437 & \underline{0.2690}\\
        \midrule

        STL REG  RAC & G& 0.8640 & 0.0217 & 0.8851 & 0.3949\\
         MTL  RAC &  & \textbf{0.8087} & \textbf{0.0141} & 0.8898 & \textbf{0.4815}\\
        \midrule

        STL REG  RAC & RN & 0.8197 & 0.0129 & 0.8972 & 0.0986\\
         MTL  RAC &  & \textbf{0.7697} & \textbf{0.01149} & 0.9012 & \textbf{0.1082}\\
        \midrule
        
        STL REG  RAC & R2CPP & 0.6130 & 0.0272 & 0.7989 & 0.0150\\
         MTL  RAC &  & \textbf{0.5157} & \underline{0.0418} & 0.7892 & \textbf{0.2486}\\    
        \bottomrule
    \end{tabular}
    } 
    \caption{Comparison of regression uncertainty method performance between STL REG and MTL models for classification tasks. The best model for each metric is in bold, while models with worse metrics than the baseline are underlined.}
    \label{tab:regression_cp_1}
\end{table}

\subsection{RQ3: Benefits of an Auxiliary Task Versus Independent Models in Collaborative Content Moderation}\label{rq3_results}

In this section, we present the main results of our collaborative content moderation framework, focusing on the outcomes of our CARE metric, F1 Review Efficiency, and Pearson correlation \(r_{I,d}\). The results are categorized by regression loss, comparing the MTL approach against the CoM approach, where STL REG and STL CLASS work together to make a final decision.

\begin{itemize}[leftmargin=*]
    
    \item MTL MSE: In Table \ref{tab:final_metric_weigthed_MSE}, we observe a clear improvement in F1 Review Efficiency and point-biserial correlation. Additionally, the FPR shows overall improvement, except for the CCLAC using the AR uncertainty method.

    \item MTL BCE: In Table \ref{tab:final_metric_weigthed_MSE}, the BCE version shows clear improvement across all metrics, except for the CCLAC using the RN uncertainty method.

    \item MTL RAC: In Table \ref{tab:final_metric_weigthed_MSE}, we observe better F1 Review Efficiency compared to the CoM models, primarily due to the improvement in the MURE metric achieved by incorporating the auxiliary task. For FPR, improvements are seen only when using the R2CCP and RN uncertainty methods, while the other cases show worse results. Regarding \(r_{I,d}\), we see overall improvements, except when using the R2CCP uncertainty method.

\end{itemize}

To sum up, the comparison shows that using a multitask approach in our proposed moderation framework leads to improvements in F1 Review Efficiency, CARE, and \(r_{I,d}\), particularly for the MSE and BCE models. While there are instances where the CoM version performs better, such as with the RAC model, the overall superiority of MTL model -- demonstrated by higher F1 Review Efficiency and greater computational efficiency -- outweighs these exceptions.

Finally, in Table \ref{tab:final_metric_baseline} we can see that the STL CLASS approach can also detect some instances of ambiguous comments as part of its uncertainty quantification process. Although its ability is limited, since it is not explicitly trained to predict annotation disagreement, the method still offers valuable insights into ambiguous cases.
\begin{table}
    \centering
    \resizebox{\textwidth}{!}{
    \begin{tabular}{cccccc}
        \toprule
        REG Method & Class CP & Regre CP & \makecell{R-F1 \\ (MTL/STL)\(\uparrow\)} & \makecell{FPR\\(MTL/STL)\(\downarrow\)} & \makecell{\( r_{pb}\) \\ (MTL/STL) \(\uparrow\)}\\
        \midrule
        
         & LAC &  & \textbf{0.553}/0.551 & \textbf{0.562}/0.564 & \textbf{0.462}/0.457 \\
         MSE & CCLAC & AR & \textbf{0.509}/0.498 & 0.563/\textbf{0.561} & \textbf{0.462}/0.454 \\
        & CRC & & \textbf{0.553}/0.551 & \textbf{0.562}/0.564 & \textbf{0.462}/0.457 \\
        \midrule
         & LAC &  & \textbf{0.553}/0.551 & \textbf{0.573}/0.596 & \textbf{0.463}/0.459 \\
         MSE & CCLAC & G& \textbf{0.509}/0.498 & \textbf{0.577}/0.612 & \textbf{0.464}/0.462 \\
        & CRC & & \textbf{0.553}/0.551 & \textbf{0.573}/0.593 & \textbf{0.463}/0.459 \\
        \midrule
         & LAC &  & 0.549/\textbf{0.555} & \textbf{0.647}/0.660 & \textbf{0.344}/0.342 \\
         MSE & CCLAC & RN & \textbf{0.509}/0.498 & 0.647/0.647 & \textbf{0.349}/0.344 \\
        & CRC & & \textbf{0.553}/0.551 & \textbf{0.647}/0.660 & \textbf{0.344}/0.342 \\
         \midrule
         & LAC &  & 0.551/0.551 & \textbf{0.562}/0.564 & \textbf{0.462}/0.457 \\
         BCE & CCLAC & AR & \textbf{0.499}/0.498 & \textbf{0.555}/0.560 & \textbf{0.465}/0.455 \\
        & CRC & & 0.551/0.551 & \textbf{0.555}/0.565 & \textbf{0.466}/0.456 \\
        \midrule
         & LAC &  & 0.551/0.551 & \textbf{0.573}/0.596 & \textbf{0.467}/0.46 \\
         BCE & CCLAC & G & \textbf{0.499}/0.498 & \textbf{0.579}/0.613 & \textbf{0.467}/0.463 \\
        & CRC & & 0.551/0.551 & \textbf{0.573}/0.596 & \textbf{0.467}/0.460 \\
        \midrule
         & LAC &  & 0.551/0.5551 & \textbf{0.653}/0.661 & \textbf{0.347}/0.340 \\
         BCE & CCLAC & RN & \textbf{0.499}/0.498 & 0.651/\textbf{0.646} & \textbf{0.353}/0.342 \\
        & CRC & & 0.551/\textbf{0.555 }& \textbf{0.647}/0.660 & \textbf{0.344}/0.342 \\
        \midrule
         
         & LAC &  & \textbf{0.582}/0.570 & 0.801/\textbf{0.762} & \textbf{0.408}/0.355 \\
         RAC & CCLAC & AR & \textbf{0.527}/0.514 & 0.801/\textbf{0.645} & \textbf{0.408}/0.373 \\
        & CRC & & \textbf{0.582}/0.57 & 0.801/\textbf{0.762} & \textbf{0.408}/0.355 \\
        
        \midrule
         & LAC &  & \textbf{0.582}/0.570 & 0.801/\textbf{0.762} & \textbf{0.408}/0.355 \\
         RAC & CCLAC & G& \textbf{0.527}/0.514 & 0.801/\textbf{0.645} & \textbf{0.408}/0.373 \\
        & CRC & & \textbf{0.582}/0.57 & 0.801/\textbf{0.762} & \textbf{0.408}/0.355 \\
        \midrule
         & LAC &  & \textbf{0.582}/0.570 & \textbf{0.427}/0.794 & 0.196/\textbf{0.344} \\
         RAC & CCLAC & R2CCP & \textbf{0.527}/0.514 & \textbf{0.403}/0.792 & 0.152/\textbf{0.352} \\
        & CRC & & \textbf{0.582}/0.570 & \textbf{0.427}/0.794 & 0.196/\textbf{0.344} \\
        \midrule
         & LAC &  & \textbf{0.582}/0.570 & \textbf{0.789}/0.845 & \textbf{0.334}/0.252 \\
         RAC & CCLAC & RN & \textbf{0.527}/0.514 & \textbf{0.789}/0.842 & \textbf{0.335}/0.259 \\
        & CRC & & \textbf{0.582}/0.570 & \textbf{0.789}/0.845 & \textbf{0.310}/0.08 \\   
        \bottomrule
    \end{tabular}
    }
    \caption{Comparison of Performance in the Moderation Review Process between MTL and CoM models. Best values for all metrics are highlighted in bold.}
\label{tab:final_metric_weigthed_MSE}
\end{table}

\begin{table}
    \centering
    \begin{tabular}{cccccc}
        \toprule
        Name & Class CP & \makecell{R-F1 \(\uparrow\)} & \makecell{FPR\(\downarrow\)} & MURE \(\uparrow\)&\makecell{\(r_{pb}\) \(\uparrow\)}\\
        \midrule
          & LAC & 0.422 & 0.237 & 0.432 &  0.258 \\
         STL CLASS & CCLAC & 0.4 & 0.281 & 0.451 & 0.258 \\
         & CRC & 0.422 & 0.237 & 0.432 &  0.258 \\
        \bottomrule        
    \end{tabular}
    \caption{STL CLASS performance on the estimation of the annotation disagreement and the moderation review process.}
\label{tab:final_metric_baseline}
\end{table}

\section{Conclusions and Future Work}\label{conclusion}


In this work, we introduce a novel framework for collaborative content moderation that addresses annotation disagreement and accounts for model uncertainty using a multitask neural network architecture, with the latter included as an auxiliary task. This approach excels at capturing the inherent subjectivity of toxic comments, providing a moderation scheme that reflects this intrinsic complexity. Additionally, we proposed two new metrics to evaluate the comment review process: \textit{F1 Review Efficiency} and \textit{CARE}. These metrics incorporate the concept of annotation disagreement into the review process, which the state-of-the-art MURE score does not consider.

Throughout the manuscript we have analysed the impact of incorporating this auxiliary task into the multitask neural architecture in terms of calibration, model performance, and uncertainty estimation for both the primary task (classification) and the auxiliary task (regression), using various regression losses and uncertainty quantification techniques. We have found that the integration of the auxiliary task improves not only the classification task, but also enhances the auxiliary task in terms of calibration and performance.

By demonstrating improvements in uncertainty estimation through the use of annotation disagreement, we have opened up an interesting new area of research, where future studies could build on towards enhancing existing classification-based tasks with better uncertainty estimation techniques. Finally, we have evaluated the quantitative benefits of employing a multitask learning architecture over a single-task neural network, obtaining promising results. The multitask approach outperformed the single-task model in detecting incorrect predictions, and also quantified disagreement in comments as specified by the chosen 
\(\gamma\), namely, the targeted ambiguity threshold.

\paragraph{Findings}We summarize the key findings for each of the research questions that have guided this study:
\begin{itemize}[leftmargin=*]
    \item RQ1: We have observed that, consistently with previous studies, the incorporation of an auxiliary task focused on annotation disagreement enhances the calibration and performance of the toxicity classification task. While it is true that overall calibration and performance improved with RAC, the gain observed was limited. In the case of the improvements of performance in the auxiliary task, an overall improvement was observed for the MSE and MAE metrics. 

    \item RQ2: In terms of improving uncertainty estimation by adding an auxiliary task, this addition has been shown to enhance the classifier’s uncertainty estimation compared to a single-task neural network. Furthermore, we observed a better point-biserial correlation between the model's uncertainty and annotation disagreement, indicating a stronger understanding of borderline cases. For the regression task (disagreement estimation), the confidence intervals generated were wider than those of the single-task regression model used for the same task, particularly for the RAC and BCE models. However, the MSE model did not demonstrate consistent performance differences, making the overall enhancement inconclusive and highly dependent on the appropriate configuration based on the selection of the uncertainty method and the regression loss.

    \item RQ3: The multitask architecture generally outperforms CoM, which consist of two separate, single-task neural networks for classification and regression. Although in some cases the improvement reported in our results is small, the MTL framework is more computationally efficient and offered advantages regarding the responses to RQ1 and RQ2. Compared to the single-task neural network, the multitask architecture provides more flexibility for moderation, as it directly computes annotation ambiguity. However, we acknowledge that the uncertainty estimated for the single-task classifier is still reasonably aligned with comment ambiguity.
\end{itemize}

On an overall concluding note, the moderation framework is not only aligned better with the inherent limitations of toxicity classification, but also offers improved uncertainty quantification, model calibration, and performance, which are beneficial for real-world moderation processes.

\paragraph{Limitations}Toxicity classification, and by extension content moderation, faces a significant challenge due to the inherent subjectivity of the task. This subjectivity complicates dataset creation and the annotation process, as we must account for both  explicit and implicit (\emph{contextualized}) toxicity. Since a few datasets provide prior comments or context for toxic comments, it becomes difficult to accurately assess the toxicity during annotation. This can lead to several issues in the annotation process, as outlined by \citet{zhang2023taxonomy}. The main challenges include: rater heterogeneity, individual differences, variations in working patterns, difficulties in understanding comment complexity, unclear task descriptions, and issues related to randomness. These challenges directly impact our proposed moderation framework, which leverages annotation disagreement as a core element. Another limitation arises from the availability of high-quality annotations. Many datasets do not ensure a sufficient number of annotators per instance, often due to economic constraints, as increasing the number of annotators significantly raises annotation costs. Since our framework depends on capturing disagreement effectively, datasets with limited annotations may not provide enough perspectives to reveal meaningful variations in subjective judgments. Additionally, our use of a balanced toxic dataset means that we have not validated how our framework performs in other settings, such as imbalanced distributions of toxic and non-toxic comments, which are more representative of real-world moderation tasks. A key limitation lies in the distribution of annotation disagreements within existing datasets. Many datasets, like Wiki Detox, are dominated by comments with minimal disagreement, meaning that annotators rarely differ significantly in their assessments. However, our approach is designed to leverage situations where there is a high level of disagreement between annotators. Therefore, the full potential of our framework is best realized when applied to datasets that capture a broader range of annotator perspectives. That being said, even in cases with little to no disagreement, our method still functions effectively as a traditional toxicity detector learned from instances with majority-voted labels.

\paragraph{Future Work} As part of our future work, we intend to explore additional uncertainty estimation techniques to investigate whether the performance gains from introducing the auxiliary task extend to other methods, such as Evidential Learning, Epistemic Neural Networks, and Bayesian Deep Learning. This will help assess the robustness and generalizability of our framework when incorporating uncertainty estimation techniques suitable for deep neural networks. Furthermore, we aim to evaluate its applicability to other tasks characterized by subjective annotations, such as sentiment analysis, where annotation disagreement and ambiguity play a critical role. Moreover, we have observed that the slight class imbalance has been counteracted by the use of focal loss. In the case of a higher degree of imbalance, a potential avenue for future work could involve incorporating other techniques to address data imbalance, such as oversampling the minority class, undersampling the majority class, and using cost-sensitive loss functions, among others. Additionally, given the broad applicability of our proposal, we plan to further test it in real-world environments and are eager to collaborate with industry leaders. This collaboration will help us refine and adapt our approach based on practical feedback, ensuring its effectiveness and relevance in real-world moderation workflows. Finally, we envision that our approach holds potential in the crossroads of active learning and generative text modelling, as it can generate more diverse datasets and eventually, improve the detection of evolving patterns in toxic language expression. In this envisioned approach, we will investigate how to effectively avoid known issues with synthetic data generation in fine-tuning loops, including the model degradation due to content autophagy \cite{xing2024ai}.

\section*{Acknowledgments}

J. Del Ser acknowledges funding support from the Basque Government through EMAITEK/ELKARTEK grants (IKUN, KK-2024/00064, BEREZ-IA, KK-2023/00012), as well as the consolidated research group MATHMODE (IT1456-22).
Guillermo Villate-Castillo acknowledges the funding support from TECNALIA Research \& Innovation, provided to employees pursuing their doctoral thesis.

\bibliographystyle{elsarticle-num-names}
\bibliography{elsarticle_bib}



\end{document}